\documentclass[10pt,twocolumn,letterpaper]{article}

\usepackage{cvpr}              
\usepackage[accsupp]{axessibility}

\usepackage[dvipsnames]{xcolor}

\definecolor{cvprblue}{rgb}{0.21,0.49,0.74}
\usepackage[pagebackref,breaklinks,colorlinks,citecolor=cvprblue]{hyperref}

\def\paperID{26} 
\def\confName{CVPR}
\def\confYear{2024}

\title{Extending global-local view alignment for self-supervised learning with remote sensing imagery}

\author{Xinye Wanyan\\
{\tt\small x.wanyan@unimelb.edu.au}
\and
Sachith Seneviratne\\
{\tt\small sachith.seneviratne@unimelb.edu.au}
\and
Shuchang Shen \\
{\tt\small chuchangs@student.unimelb.edu.au}
\and
Michael Kirley \\
{\tt\small mkirley@unimelb.edu.au}
}

\begin{document}
\maketitle
\begin{abstract}
Since large number of high-quality remote sensing images are readily accessible, exploiting the corpus of images with less manual annotation draws increasing attention.
Self-supervised models acquire general feature representations by formulating a pretext task that generates pseudo-labels for massive unlabeled data to provide supervision for training.
While prior studies have explored multiple self-supervised learning techniques in remote sensing domain, pretext tasks based on local-global view alignment remain underexplored, despite achieving state-of-the-art results on natural imagery.
Inspired by DINO \cite{caron2021emerging}, which employs an effective representation learning structure with knowledge distillation based on global-local view alignment, we formulate two pretext tasks for self-supervised learning on remote sensing imagery (SSLRS).
Using these tasks, we explore the effectiveness of positive temporal contrast as well as multi-sized views on SSLRS. 
We extend DINO and propose DINO-MC which uses local views of various sized crops instead of a single fixed size in order to alleviate the limited variation in object size observed in remote sensing imagery.
Our experiments demonstrate that even when pre-trained on only 10\% of the dataset, DINO-MC performs on par or better than existing state-of-the-art SSLRS methods on multiple remote sensing tasks, while using less computational resources.
All codes, models, and results are released at \url{https://github.com/WennyXY/DINO-MC}.
\end{abstract}    
\section{Introduction}
\label{sec:intro}
Computer vision models provide promising solutions for remote sensing image analysis to be applied in numerous real-world tasks, including disaster prevention \cite{2000remote}, forestry \cite{2021Review}, agriculture \cite{Mulla2013Twenty}, land surface change \cite{2021A}, biodiversity \cite{turner2003remote}.
However, training large-scale deep learning models in a traditional supervised manner requires extensive labeled datasets. 
Labeling large datasets is costly and error-prone, particularly within the remote sensing domain, which requires expert knowledge \cite{stojnic2021self}.
Therefore, reducing the reliance of the model on labeled images is crucial for resolving specific downstream tasks.
Self-supervised learning (SSL) paradigm is a common solution to learn useful features from copious data without labels, which comprises two distinct phases: the initial pretext task pre-training phase followed by the subsequent downstream task fine-tuning phase.
Since different pretext tasks prompt the model to acquire diverse feature aspects, devising an appropriate task for which labels can be automatically derived from the data is essential for fostering generalized representations.

Based on the pretext tasks, SSL models can be categorized into three distinct categories.
Generative tasks allow the model learn feature representations by reconstructing or generating original images.
For example, the image inpainting \cite{pathak2016context} uses the original data as labels to train the model to recover several masked parts of the original image.
The discriminative tasks \cite{chen2020simple,dosovitskiy2014discriminative} train the network to distinguish certain characteristics or properties.
For example, \cite{doersch2015unsupervised} trains the model to predict the relative position of a patch to its neighbors, and each image patch is defined to have up to eight adjacent patches.
However, the feature representations learned by generative and discriminative methods highly rely on the designed pretext tasks, and an ineffective pretext task might reduce the transfer-ability of the pre-trained model \cite{wang2022selfreview}.
In stead of solving a straightforward and specific pretext task, contrastive SSL approaches extract features by maximising the similarity between the latent representations of two positive instances (e.g., two augmented views of the same image) and the distance between two negative instances. 
However, simply following this approach will easily lead to an identity map for each pair of positive samples, i.e., model collapse \cite{wang2022selfreview}.

DINO \cite{caron2021emerging} utilizes knowledge distillation and centering and sharpening of the teacher network \cite{hinton2015distilling} to handle this issue.
DINO exhibits impressive performance in numerous computer vision tasks, including image retrieval, copy detection, and video instance segmentation. 
Although some previous work has applied DINO to remote sensing domain \cite{wang2022last,seneviratne2021self,wang2022selfSAR}, a comprehensive evaluation and extension of the self-supervised objective for remote sensing imagery deserves further exploration. 
In particular, the size of instances observed during pre-training on traditional natural scenes shows wider variation than seen in remote sensing imagery. 
Thus, the alignment of the latent representation of the multi-sized local augmented crops against the global augmented views is of interest in remote sensing imagery as it leads to a more challenging pretext task.
Remote sensing data holds spatial-temporal heterogeneity offering rich time series information due to the repetitive capture of imagery by satellites over the same geographic area. 
Some studies \cite{bai2020can, manas2021seasonal, wang2023ssl4eo} have explored employing the temporal information for SSL and their results indicate promising potential.
\cite{bai2020can} integrated temporal information of video sequences into instance discrimination based contrastive SSL.
\cite{manas2021seasonal, wang2023ssl4eo} constructed large-scale SSLRS datasets involving a wide spatio-temporal range of remote sensing imagery.

Inspired by the inherent characteristics of the size variation of semantic content observed between traditional natural scenes and remote sensing imagery, we propose DINO-MC which uses size variation in local crops to drive better representation learning of the semantic content in remote sensing imagery.
Additionally, we explore applying temporal views as positive instances into contrastive SSL method. 
We evaluate our model with different backbones including two transformer-based models and two convnets.
Although the majority of current SSL methods for remote sensing imagery rely on ResNet and Vision Transformers (ViTs) as backbone architectures, there is growing interest in exploring the self-supervised feature extraction capabilities of Wide ResNets (WRN) \cite{zagoruyko2016wide} and Swin Transformers \cite{liu2021swin} in the remote sensing domain.
We incorporate these two architectures into our analysis.
In the linear probing evaluation, the findings demonstrate that DINO-MC has great transfer-ability which achieves 2.56\% higher accuracy with a smaller pre-trained dataset than SeCo.
When fine-tuned on two remote sensing classification tasks and change detection task, DINO-MC outperforms DINO and SeCo.

Our main contributions are summarized as follows:
\begin{itemize}
\item[$\bullet$] We apply temporal views as positive instances to recent contrastive self-supervised models (DINO-TP). We analyze different backbone networks to explore their effectiveness on different remote sensing tasks when pre-trained under this setting.
\item[$\bullet$] We combine a new multi-sized local cropping strategy with DINO and propose DINO-MC. We pre-train DINO-MC with different backbones on a satellite imagery dataset SeCo-100K to learn general representations.
\item[$\bullet$] DINO-MC outperforms SeCo with only 10\% pre-training dataset, and achieves state-of-the-art results on BigEarthNet multi-label and EuroSAT multi-class land use classification, as well as Onera Satellite Change Detection (OSCD) task.
\end{itemize}

\begin{figure*}
  \centering
  \includegraphics[width=0.8\linewidth]{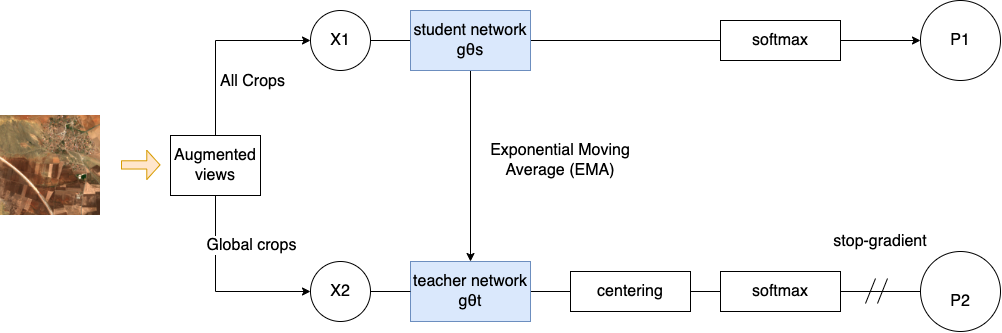}
  \caption{
  \textit{DINO}: the self-supervised contrastive algorithm with knowledge distillation.
  It is the basic structure of both DINO-TP and DINO-MC.
  For DINO-TP, we use three temporal views to generate global crops and multi-sizes local crops as the input to do positive contrastive representation learning.
  For DINO-MC, we generate global and local crops from one imagery, then apply two different augmentations to global views and multi-sizes local views, respectively, to get the input of teacher and student network.
  }
  \label{fig:dino}
\end{figure*}
\section{Related Work}
\subsection{Contrastive Self-supervised Learning.}
SSL methods create task-agnostic signals to instruct the model in learning versatile features which can effectively generalise across various downstream tasks \cite{tao2023self}.
In generative and discriminative SSL, the loss function is computed based on the disparity between the ground truth and the model's prediction. 
Conversely, in contrastive models, the loss is determined by the dissimilarities among feature representations within the latent space \cite{wang2022selfreview}.
Contrastive SSL exhibits promise in transferability and gains increasing attention.
Furthermore, the statistics of recent studies shows that nearly half of the remote sensing self-supervised studies applied contrastive learning models \cite{wang2022selfreview}.

\textbf{Instance discrimination} \cite{dosovitskiy2014discriminative} proves to be a highly effective pretext task commonly used in contrastive learning. 
In this task, augmented views originating from the same input are labeled as positive, whereas distinct instances are designated as negative.
The model is then trained to maintain proximity between positive pairs and create separation between negative pairs in the representation space \cite{wu2018unsupervised, tian2020makes,seneviratne2021self}.
Several studies have explored positive/negative samples generating strategies.
\cite{oord2018representation} regards the diagonally connected patches as the positive instances otherwise negative, and \cite{tian2020contrastive} incorporates same spot views from different sensory as the positive instances.
\cite{he2020momentum} proposes MoCo-V1 with a momentum encoder to effectively maintain multiple negative samples for representation learning.
SimCLR \cite{chen2020simple} employs very large batch sizes instead of momentum encoders, and provides experimental results on $10$ forms of data augmentation.
Inspired by SimCLR, MoCo-V1 is extended to MoCo-V2 \cite{chen2020improved}, and MoCo-V3 \cite{chen2021empirical} integrates ViT as the backbone architecture.

\textbf{Clustering method} is another line of contrastive learning, which demands no precise indication from the inputs \cite{caron2018deep}. 
\cite{caron2020unsupervised} introduces a contrasting cluster assignment that learns features by Swapping Assignments between multiple Views of the same image (SwAV).
SwAV can be readily applied to different sizes of datasets due to its utilization of clustering rather than pairwise comparisons.
\cite{caron2020unsupervised} proposes a data augmentation named multi-crop to increase the instances without drastically additional memory and computation.  
Specifically, when doing multi-crop, images are cropped to a collection of views with lower resolutions instead of the full-resolution views.
\cite{assran2022masked} maps the feature representations of two randomly augmented views (with one view having randomly masked patches) to a shared cluster space to maximize their similarity.

\textbf{Knowledge distillation} in contrastive learning can be employed for feature extraction without the need to differentiate between images \cite{caron2021emerging}.
\cite{grill2020bootstrap} proposes Bootstrap Your Own Latent (BYOL) based on the online and target networks whose weights are updated with each other.
Inspired by BYOL, \cite{caron2021emerging} explores the further synergy between SSL and different backbones, especially ViTs, and proposes a simple form of self-distillation with no labels (DINO).
DINO uses same networks architecture as BYOL but with different loss function and backbones.
When pre-trained on ImageNet \cite{russakovsky2015imagenet}, DINO achieves better linear and KNN probing evaluation results than other self-supervised methods with fewer computation resources \cite{grill2020bootstrap,caron2020unsupervised,caron2021emerging}.
\cite{Song_2023_CVPR} suggests incorporating two distillation modes within SSL: self-distillation and cross-distillation, to improve the semantic feature alignment across two networks.
\cite{Jang_2023_WACV} enables the intermediate layers to learn from the final layer by employing contrastive loss through, i.e., self-distillation.

\subsection{Self-supervised Learning in Remote Sensing.} 
\cite{tao2020remote} experiments with SSL models on different pretext tasks, including image inpainting \cite{pathak2016context}, context prediction \cite{doersch2015unsupervised}, and instance discrimination \cite{wu2018unsupervised} on remote sensing imagery tasks. 
\cite{vincenzi2021color} proposes to learn practical representations from satellite imagery by reconstructing the visible colors (RGB) from its high-dimensionality spectral bands (Spectral).
\cite{ayush2021geography} explores the application of contrastive SSL to large-scale satellite image datasets.
\cite{manas2021seasonal} leverages seasonal information of remote sensing imagery and demonstrates that the performance of MoCo-V2 is enhanced with the incorporation of temporal positives (TP).
TP regards the temporal views as the positive instances and trains models match their representations allowing the model to learn essential features that do not change over time.
Besides, they introduce a negative temporal contrastive model named SeCo designed to capture variations or discrepancies resulting from temporal changes.
\cite{wang2022last} uses Swin Transformer as the backbone of DINO and applies it to remote sensing imagery tasks.
DINO-MM \cite{wang2022selfSAR} extends DINO by combining synthetic-aperture radar (SAR) and multispectral (optical) images and is applied to BigEarthNet land use classification task.

Current studies have demonstrated the utility and viability of SSL models for practical applications in tasks involving remote sensing images. But the full potential of SSL in remote sensing warrant further investigation.
Our work targets to bridge this gap and extend existing SSL model by generating more effective contrastive instances.

\section{Methodology}
Our work is mainly based on a contrastive SSL model named DINO, which has been applied into both natural and remote sensing imagery tasks \cite{caron2021emerging,wang2022selfSAR,wang2023ssl4eo}.
We aim to learn useful, transferable features for remote sensing imagery by exploring positive temporal contrastive SSL model DINO-TP (\cref{section:dino-tp}) and contrastive SSL model with multi-sized crops DINO-MC (\cref{section:dino-mc}).
In addition, we experiment with the feature extraction ability of different backbones in SSL.

\subsection{DINO-TP}
\label{section:dino-tp}
\textbf{Architecture }
\cref{fig:dino} shows the model structure.
The student and teacher networks in DINO have same architecture $g$ but different weights $\theta_t$ and $\theta_s$.
Two sets of augmented views $x_1$ and $x_2$ are generated from the same image.
The global view covers the majority of the initial image and the local view only covers a small part.
The student network receives both global and local crops as inputs, whereas the teacher network only receives global crops.
The model is trained to match the individual local views to global views in the feature space.

Rather than a pre-trained and frozen teacher network \cite{fang2021seed,chen2020big}, DINO dynamically updates the teacher weights by using EMA on student weights, i.e.,  
\begin{equation}
  \theta_t\gets\theta_t\lambda+(1-\theta_s)
  \label{eq:teacher_w}
\end{equation}
The $\lambda$ values adhere to a cosine schedule between $0.996$ and $1$, indicating that the teacher network is less dependent on the present student and more dependent on the integration of the student network in each round; hence, the weights are updated slowly.

Centering and sharpening are used to prevent model collapse.
Centering is adding a bias term $c$ to the features of the output of the teacher model, i.e., 
\begin{equation}
  g_t(x) = g_t(x) + c
  \label{eq:centering}
\end{equation}
The bias term $c$ is dynamically updated by the EMA, where $m > 0$ denotes the update rate and $B$ is the batch size.
\begin{equation}
  c\gets mc+(1- m)\frac{1}{B}\sum_{i=1}^B g\theta_t(x_i)
  \label{eq:c_update}
\end{equation}
Sharpening is achieved by softmax normalization using low temperature in the teacher network to avoid consistent distribution.
The teacher network consistently outperforms the student network during pre-training, which is used as the feature extractor in downstream tasks \cite{caron2021emerging}.

\begin{figure*}
  \centering
  \includegraphics[width=0.8\linewidth]{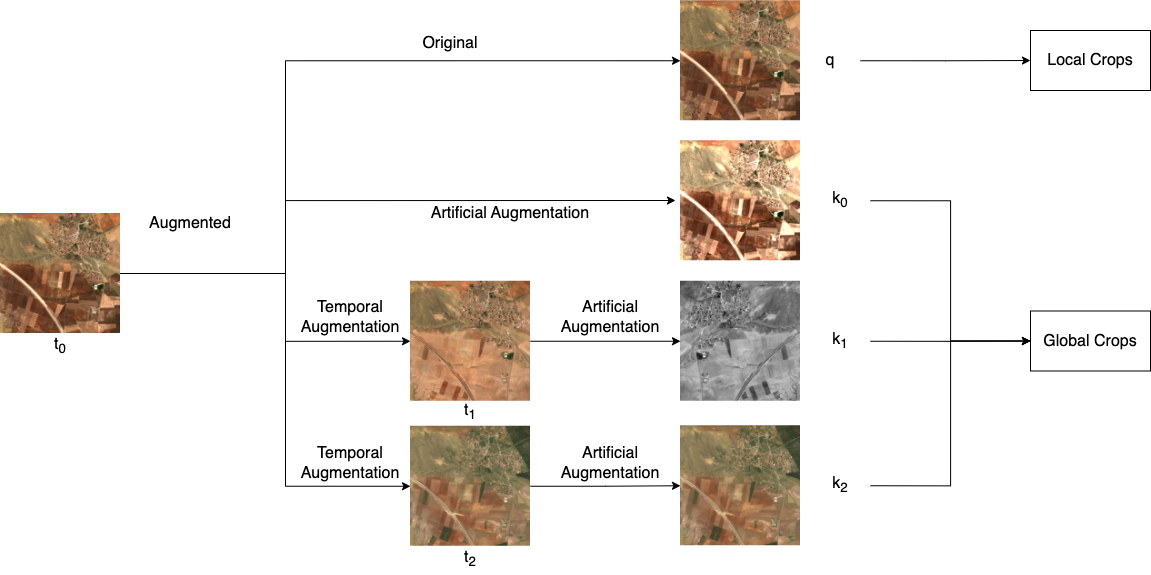}
  \caption{
  The process of handling temporal views of contrastive learning in DINO-TP.
  We randomly select three temporal views of the same location and augment them to obtain global and local crops, which is the input of the teacher and student network.
  Different temporal views of the same location in DINO-TP are considered as positive examples, and we train the model to match their representations in the feature space.
  }
  \label{fig:temporal_aug}
\end{figure*}

\textbf{Temporal Contrastive} 
This work performs self-supervised pre-training on SeCo-100K \cite{manas2021seasonal} to integrate temporal information.
SeCo-100K is an unlabeled remote sensing imagery dataset collected from Sentinel-2 \cite{drusch2012sentinel} with 100K instances, and each instance is composed of five different temporal views of the same location.
DINO-TP regards the temporal views of the same location as the positive instances, i.e., the teacher and student networks are trained to match temporal views in the feature space.
In the pre-training phase, we randomly select three temporal views and name them as $t_0, t_1, t_2$.
As shown in \cref{fig:temporal_aug}, $t_1, t_2$ can be regarded as the temporal augmentation views of $t_0$.
The temporal image $t_0$ is directly used as $q$ to generate $6$ local crops of various sizes, including $184^2, 164^2, 144^2, 124^2, 104^2$ and $84^2$.
After applying color jittering, grayscale, and Gaussian blur to each of the three temporal perspectives, we acquire $k_0, k_1, k_2$ views.
$k_0, k_1, k_2$ are scaled and resized into $224^2$ serving as three global crops utilized as the inputs.
In prior work, different techniques are paired with different crop scaling ranges.
DINO-TP not only learns the relationship between the whole and the pieces of the image, but also discovers the interrelation among various temporal perspectives.

\subsection{DINO-MC}
\label{section:dino-mc}
The structure of DINO-MC is nearly identical to DINO-TP.
This work changes the cropping strategy for local crops in DINO to create a more challenging pretext task.
We use multi-sized crops instead of the fixed-size local crops used in DINO. 
Keeping the number of global and local crops same as DINO, DINO-MC outperforms DINO on both linear and KNN probing as well as end-to-end evaluation on multiple downstream tasks.
Color-related pretext tasks have been proven to be effective in multiple image tasks in the field of remote sensing \cite{zhang2016colorful, vincenzi2021color, assran2022masked}.
Due to the strong connection between color and semantics, we set two strategies of color transformations for global and local crops respectively.
The global views are augmented by random color jittering and GaussianBlur.
The local views are augmented by random color jittering, random grayscale shifting, and random Gaussian blur all together.
Random color jittering modifies the brightness, contrast, saturation and hue of the image with a specified probability within a certain range.
This process aims to replicate the effects of capturing images in diverse lighting conditions and real-world shooting scenarios.
Random grayscale converts an image into a grayscale image randomly, aiming to diminish the influence of color while capturing additional image characteristics beyond color properties.
We use different settings for cropping and color transformation.
The experiments verify that our strategy is effective and DINO-MC outperforms DINO on both linear probing and end-to-end fine-tuning on three downstream tasks (classification and change detection).

\section{Experiments}
In this study, the features learned from the self-supervised pre-training are evaluated on three downstream tasks: EuroSAT \cite{helber2019eurosat} and BigEarthNet \cite{sumbul2019bigearthnet} land use classification tasks, and OSCD change detection task \cite{daudt2018urban}.
Due to its flexibility and computational efficiency, we utilize DINO as the foundation for our experiments and further extend its capabilities.
SeCo-100K \cite{manas2021seasonal} serves as the dataset for self-supervised representation learning in this experiment, containing 100K unlabeled remote sensing images.
During the pre-training phase, the student and teacher networks receive two separate sets of temporal views depicting the same geographical region. 
The SSL model is trained to aligns these views within the feature space to generate representations that exhibit temporal consistency.
In DINO-MC and DINO-TP, the scaling range of multi-crop is $(0.05, 0.32)$ for local crops and $(0.32, 1)$ for global crops.

\subsection{DINO with Different Backbones}
When employed as the backbone of DINO, ViT demonstrates superiority over ResNet \cite{caron2021emerging}.
In this work, there are four different networks applied as the backbone of DINO, DINO-TP, and DINO-MC, including ViT, Swin Transformer, ResNet, and WRN.

ViT preserves the Transformer structure used in NLP as much as possible and exhibits promising performance in computer vision tasks \cite{dosovitskiy2020image}.
The input of transformer encoder is the embedding formed by adding the patch embedding with the position encoding.
The transformer encoder of ViT consists mainly of layer normalization (LN) which is applied before each block, multi-head attention, and multi-layer perceptron block (MLP).
Inspired by ViT, Swin Transformer is proposed with a hierarchical transformer, which computes representation with both regular and shifted windows to capture features from different levels and resolutions \cite{liu2021swin}. 

A deep residual learning framework (ResNet) is proposed to overcome the degradation issue of the deep neural networks \cite{he2016deep}.
They add shortcut connections to transmit the information of shallow layers directly to the deeper layers of the neural network and require no additional computation.
With this mechanism, the residual net is deepened to $152$ layers and achieves state-of-the-art results in multiple computer vision tasks. 
Furthermore, with the emerging and development of self-supervised representation learning, ResNet also serves as an effective backbone widely used in multiple SSL architectures \cite{chen2020simple, grill2020bootstrap, oord2018representation, tian2020contrastive, misra2020self}. 
WRN \cite{zagoruyko2016wide} is proposed to improve the performance of ResNet by increasing the width of the residual block, i.e., widening the convolutional layers.
When employing an equal number of parameters, wide residual blocks yield superior results with reduced training time compared to the original residual blocks.
Despite the quadratic increase in parameter numbers and computational complexity associated with the width factor, it proves to be a more efficient approach compared to expanding the number of layers in ResNet. This is because larger tensors can better leverage the parallel computing capabilities of GPUs \cite{zagoruyko2016wide}.
WRN has been proved to achieve advanced results in a supervised manner in terms of classification F1-Score and training efficiency \cite{papoutsis2021efficient}.
This project intends to explore the feature extraction capability of WRN in SSL framework.

\textbf{Implementation Details }
ViT-small is used as the backbone and the implementation follows DINO.
In the experiments, we load different backbones as the teacher and student network without pre-trained weights.
WRN-50-2 has similar structure to ResNet, with the exception of the bottleneck number of channels, which is twice as large in each block.
We follow the recommendation from DINO to use AdamW optimizer.
Cross-entropy is employed as the loss function to measure the disparity between feature representations generated by two networks.
We evaluate the learned representations by applying KNN and linear probing on EuroSAT land use classification task.

\textbf{Quantitative Results}
A common method for assessing feature representations is training a classifier on top of the frozen feature extraction model for a supervised learning assignment.
In order to evaluate the representations independently, we freeze the pre-trained backbone model and only train the linear and KNN classifiers on EuroSAT land use classification task.
The results are shown in \cref{table:DINO-Backbones}.
From the table, DINO-MC performs better than both DINO and DINO-TP when using ViT-samll, WRN-50-2, and ResNet-50 backbones.
DINO-MC with WRN-50-2 pre-trained on 100K data is even 2.56\% higher than the linear probing accuracy of SeCo pre-trained on 1 million data.
Additionally, DINO-MC with ViT-samll has 2.59\% higher accuracy than DINO with ViT-samll which indicates the effectiveness of our strategy.
In comparison to the other three backbones, Swin-tiny performs less well.
One possible reason could be that the model size of Swin-tiny is much smaller than the other three models.
Another interesting observation is that the DINO-TP performs worse with two convnets than the DINO, but better with two transformer models.
Among the different self-supervised models, ViT-small and Swin-tiny performed more consistently than ResNet-50 and WRN-50-2.
Overall, DINO-MC is particularly effective.

\begin{table}
  \centering
  \begin{tabular}{lcccc}
    \toprule
    Model & Arch & \#images & KNN & Linear \\
    \midrule
    MoCo-V2 & ResNet-50 & 1M & - & 83.72 \\
    SeCo-1M & ResNet-50 & 1M & - & 93.14 \\
    \midrule
    DINO & ResNet-50 & 100K & 90.09 & 89.65  \\
    DINO-TP & ResNet-50 & 100K & 79.05 & 86.70  \\
    DINO-MC & ResNet-50 & 100K & 93.94 & 95.59  \\
    \midrule
    DINO & WRN-50-2 & 100K & 92.74 & 91.65 \\
    DINO-TP & WRN-50-2 & 100K & 86.37 & 88.15 \\
    DINO-MC & WRN-50-2 & 100K & \textbf{94.65} & \textbf{95.70} \\
    \midrule
    DINO & ViT-small & 100K & 93.35 & 91.50 \\
    DINO-TP & ViT-small & 100K & 93.15 & 93.89 \\
    DINO-MC & ViT-small & 100K & 93.41 & 94.09 \\
    \midrule
    DINO & Swin-tiny & 100K & 92.15 & 86.87 \\
    DINO-TP & Swin-tiny & 100K & 92.83 & 91.94 \\
    DINO-MC & Swin-tiny & 100K & 93.22 & 90.54 \\
    \bottomrule
  \end{tabular}
  \caption{Linear and KNN probing classification on EuroSAT. 
  We evaluate our models with different backbones on EuroSAT and record the top-1 accuracy of KNN and linear probing on the validation set.
  MoCo-V2 \cite{chen2020improved} and SeCo-1M \cite{manas2021seasonal} are pre-trained on SeCo-1M dataset, and their linear probing results on EuroSAT are from SeCo \cite{manas2021seasonal}.
  Other listed models are pre-trained on SeCo-100K dataset with only 10\% images of SeCo-1M.
  }
  \label{table:DINO-Backbones}
\end{table}

\subsection{Land Use Classification on EuroSAT}
EuroSAT is a widely used benchmark dataset for remote sensing land use classification tasks.
It is used for representation evaluation in this study, allowing the results of models based on it to be easily compared to those of other models.
The dataset, which collects 27,000 remote sensing images from the Sentinel-2 satellite, is divided into 21,600 and 5,400 images for training and evaluation respectively in our experiments.

\textbf{Implementation Details }
The pre-trained backbones of the SSL models are evaluated as feature extractors in this land use classification downstream task.
Based on the pre-trained backbone models, we incorporate a fully-connected layer as the classifier, leveraging the extracted features to predict the classification results.
Both the feature extractor and the classifier are fine-tuned on this supervised classification task.
We train the models for around 200 epochs with a batch size of 32.
The learning rate is $1e-3$ or $3e-3$ for different models.
We use the \textit{CosineAnnealingLR} scheduler to update the learning rate.
Same as SeCo \cite{manas2021seasonal}, we use SGD optimizer without weight decay to update weights.

\textbf{Quantitative Results}
\cref{table:EuroSAT_results} compares our pre-trained models against three supervised baseline models on EuroSAT land use classification task.
DINO-MC with WRN-50-2 achieves similar results to the three supervised models, in particular, it is pre-trained on only 100K images, while the three supervised models are pre-trained on 1M images.
This further confirms the effectiveness of the representations learned by DINO-MC.

\begin{table}
  \centering
  \begin{tabular}{lcc}
    \toprule
    Model & Backbone & Accuracy \\
    \midrule
    Supervised & WRN-50-2 & 98.72 \\
    Supervised & ResNet-50 & 98.78 \\
    Supervised & ViT-base & 98.83 \\
    \midrule
    DINO & ViT-small & 97.98 \\
    DINO-MC  & ViT-small & 98.15 \\
    DINO-MC  & Swin-tiny & 98.43 \\
    DINO-MC  & ResNet-50 & 98.69 \\
    DINO-MC  & WRN-50-2 & \textbf{98.78} \\
    \bottomrule
  \end{tabular}
  \caption{
  End-to-end accuracy results on EuroSAT land use classification task.
  The three supervised models listed are pre-trained on ImageNet-1K \cite{russakovsky2015imagenet}, and we load and fine-tune them on EuroSAT.
  While DINO and DINO-MC are only pre-trained on SeCo-100K dataset, which has only 10\% the number of images of ImageNet-1K.
  }
  \label{table:EuroSAT_results}
\end{table}

\subsection{Land Use Classification on BigEarthNet}
BigEarthNet \cite{sumbul2019bigearthnet} is a widely used benchmark dataset for land use classification task, containing a total of 590,326 images.
This work uses BigEarthNet-S2 \cite{sumbul2019bigearthnet}, which specifically gathers remote sensing images from Sentinel-2 only, and each image is annotated by multiple land use categories.
The dataset provides 12 spectral bands for each image and a JSON file with its multi-labels and metadata information.
Following SeCo \cite{manas2021seasonal}, we employ a new nomenclature of 19 classes introduced in \cite{sumbul2020bigearthnet}, and around $12\%$ of the patches that are totally masked by seasonal snow, clouds, or cloud shadows are eliminated in this experiment.
We used the training/validation splitting strategy suggested by \cite{neumann2019domain} with 311,667 instances for training and 103,944 images for validation.

\textbf{Implementation Details }
We add a linear classification layer on top of the backbone model as the output layer and then fine-tune it on 10\% and 100\% BigEarthNet, respectively, to evaluate the features learned by the SSL models.
In this experiment, the Adam and AdamW optimizer with default hyper-parameters are used to update the weights of the models. 
Identical to SeCo, we set the learning rate to $1e-5$ and scale it down by ten in epochs of 60\% and 80\%, respectively.
We use the MultiLabelSoftMarginLoss as the loss function, which allows assigning a different number of target classes to each sample.

\textbf{Quantitative Results}
\cref{table:BigEarthNet_results} provides the results of pre-trained models in the end-to-end BigEarthNet classification task.
We assess the performance of each model by mean average precision (MAP).
When fine-tuned on the 10\% BigEarthNet, DINO-MC outperforms SeCo-100K with the same backbone.
Additionally, DINO-MC with ViT-small achieves 1.65\% higher MAP than with ResNet-50, 2.48\% higher MAP than SeCo-100K, and even 1.58\% higher MAP than SeCo-1M.
The performance of DINO-MC with each of the four backbone models exceeds that of SeCo-100K, and even outperforms SeCo-1M except for ResNet-50.

When fine-tuned on the whole BigEarthNet, DINO-MC achieves comparable result to SeCo-100K with the same backbone.
Interestingly, DINO-MC with Swin-tiny achieves 1.89\% higher MAP than with ResNet-50, 1.63\% higher MAP than SeCo-100K, and 0.94\% higher than SeCo-1M.
The results demonstrate that the representations learned by DINO-MC generalize well in the multi-label classification task on BigEarthNet.

\begin{table}
  \centering
  \begin{tabular}{lcccc}
    \toprule
    Model & Backbone & Param. & 10\% & 100\% \\
    \midrule
    SeCo-100K & ResNet-50 & 23M & 81.72 & 87.12 \\
    SeCo-1M & ResNet-50 & 23M & 82.62 & 87.81 \\
    \midrule
    DINO & ResNet-50 & 23M & 79.67 & 85.38 \\
    DINO-TP  & ResNet-50 & 23M & 80.10 & 85.20 \\
    DINO-MC  & ResNet-50 & 23M & 82.55 & 86.86 \\
    DINO-MC  & WRN-50-2 & 69M & 82.67 & 87.22 \\
    DINO-MC & Swin-tiny & 28M & 83.84 & \textbf{88.75} \\
    DINO-MC  & ViT-small & 21M & \textbf{84.20} & 88.69 \\
    \bottomrule
  \end{tabular}
  \caption{Mean average precision (MAP) results on BigEarthNet-S2 land use classification. 
  We use the same train/validation splits as SeCo \cite{manas2021seasonal}.
  We fine-tune the pre-trained SSL models with different backbones on 10\% and 100\% training dataset, and evaluate them on the identical validation dataset.
  SeCo-100K and SeCo-1M represent SeCo pre-trained on SeCo-100K and SeCo-1M, respectively, and their results listed are from \cite{manas2021seasonal}.
  DINO, DINO-MC, and DINO-TP are pre-trained on SeCo-100K only.
  }
  \label{table:BigEarthNet_results}
\end{table}

\subsection{Change Detection on OSCD}

\begin{table}
  \centering
  \begin{tabular}{lcccc}
    \toprule
    Model & Backbone & Pre. & Rec. & F1 \\
    \midrule
    Supervised & ResNet-50 & 56.49 & 43.63 & 48.61 \\
    Supervised & WRN-50-2 & 53.76 & 47.11 & 49.72 \\
    \multicolumn{2}{l}{PatchSSL (21M param.)} & 40.44 & 69.10 & 51.00 \\
    MoCo-V2 & ResNet-50 & 64.49 & 30.94 & 40.71 \\
    SeCo-1M & ResNet-50 & 65.47 & 38.06 & 46.94 \\
    \midrule
    DINO & ResNet-50 & 57.37 & 44.21 & 49.53 \\
    DINO-MC & ResNet-50 & 51.94 & 54.04 & 52.46 \\
    DINO-TP & ResNet-50 & 51.10 & 49.03 & 49.74 \\
    \midrule
    DINO & WRN-50-2 & 53.58 & 52.28 & 52.41 \\
    DINO-MC & WRN-50-2 & 49.99 & 56.81 & \textbf{52.70} \\
    DINO-TP & WRN-50-2 & 55.77 & 47.30 & 50.61 \\
    \bottomrule
  \end{tabular}
  \caption{Fine-tuning accuracy results on OSCD change detection task.
  We adopt the same train/validation splits as SeCo \cite{manas2021seasonal}.
  Following SeCo, we freeze the pre-trained backbone and only update the weights of U-net \cite{ronneberger2015u}.
  Our WRN-50-2 and ResNet-50 models are pre-trained on 100K satellite images.
  The result of PatchSSL is from \cite{chen2021self}.
  The listed MoCo-V2 \cite{chen2020improved} and SeCo-1M are pre-trained on 1M satellite images and are results provided by \cite{manas2021seasonal}.
  }
  \label{table:OSCD_results}
\end{table}

\begin{figure*}
  \centering
  
  \begin{subfigure}{0.138\linewidth}
    \centering
    \includegraphics[width=1\textwidth]{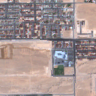}
    \subcaption*{
    Image 1 \protect \\
    (Lasvegas)
    }
  \end{subfigure}
  \begin{subfigure}{0.138\linewidth}
    \centering
    \includegraphics[width=1\textwidth]{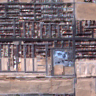}
    \subcaption*{Image 2 \\
    (Lasvegas)}
  \end{subfigure}
  \begin{subfigure}{0.138\linewidth}
    \centering
    \includegraphics[width=1\textwidth]{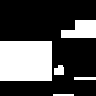}
    \subcaption*{Mask \\
    (Ground truth)}
  \end{subfigure}
  \begin{subfigure}{0.138\linewidth}
    \centering
    \includegraphics[width=1\textwidth]{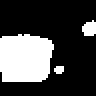}
    \subcaption*{SeCo-1M \\
    F1=84.25}
  \end{subfigure}
  \begin{subfigure}{0.138\linewidth}
    \centering
    \includegraphics[width=1\textwidth]{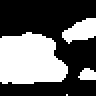}
    \subcaption*{DINO \\
    F1=86.83}
  \end{subfigure}
  \begin{subfigure}{0.138\linewidth}
    \centering
    \includegraphics[width=1\textwidth]{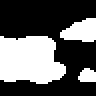}
    \subcaption*{DINO-TP \\
    F1=88.60}
  \end{subfigure}
  \begin{subfigure}{0.138\linewidth}
    \centering
    \includegraphics[width=1\textwidth]{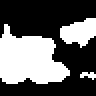}
    \subcaption*{
    DINO-MC \\
    F1=86.69
    }
  \end{subfigure}

\hspace{5mm}

    \begin{subfigure}{0.138\linewidth}
    \centering
    \includegraphics[width=1\textwidth]{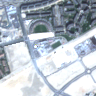}
    \subcaption*{Image 1 \\
    (Dubai)}
  \end{subfigure}
  \begin{subfigure}{0.138\linewidth}
    \centering
    \includegraphics[width=1\textwidth]{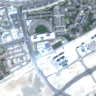}
    \subcaption*{Image 2 \\
    (Dubai)}
  \end{subfigure}
  \begin{subfigure}{0.138\linewidth}
    \centering
    \includegraphics[width=1\textwidth]{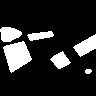}
    \subcaption*{Mask \\
    (Ground truth)}
  \end{subfigure}
  \begin{subfigure}{0.138\linewidth}
    \centering
    \includegraphics[width=1\textwidth]{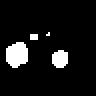}
    \subcaption*{SeCo-1M \\ 
    F1=61.89}
  \end{subfigure}
  \begin{subfigure}{0.138\linewidth}
    \centering
    \includegraphics[width=1\textwidth]{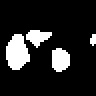}
    \subcaption*{DINO \\ 
    F1=65.40}
  \end{subfigure}
  \begin{subfigure}{0.138\linewidth}
    \centering
    \includegraphics[width=1\textwidth]{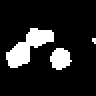}
    \subcaption*{DINO-TP \\ 
    F1=59.64}
  \end{subfigure}
  \begin{subfigure}{0.138\linewidth}
    \centering
    \includegraphics[width=1\textwidth]{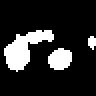}
    \subcaption*{ DINO-MC\\
     F1=68.42}
  \end{subfigure}
  \caption{
  Same as SeCo \cite{manas2021seasonal}, our visualization is based on two instances of 'Losvegas' and 'Dubai' from the OSCD change detection dataset. SeCo-1M is SeCo model pre-trained on SeCo-1M dataset, while DINO, DINO-TP, and DINO-MC are pre-trained on only SeCo-100K.
  We visualize the outputs of DINO-TP and DINO-MC for comparison with DINO and SeCo.  
  We present results on the same images as SeCo for comparison and the two outputs of SeCo is from \cite{manas2021seasonal}.
  We also provide F1 score for each output on the bottom of output image.}
  \label{Figure:OSCD_visualization}
\end{figure*}

OSCD is a benchmark dataset of which the images are collected from Sentinel-2, which mainly focuses on urban growth and disregards natural changes \cite{daudt2018urban}. 
Change detection is a fundamental problem in the field of earth observation image analysis.
The input is a pair of images captured at the same location, while the label is a mask map highlighting the change parts.
It is a binary classification task in which labels are assigned to each pixel based on a series of images sampled at different times: change (positive) or no change (negative).
The results are measured by F1 Score, precision, and recall in this experiments.

\textbf{Implementation Details }
The OSCD dataset is divided into 14 and 10 pairs for training and validation separately, as recommended by prior research \cite{daudt2018urban, manas2021seasonal}.
In addition, the categories of change and unchanged in this dataset are unbalanced due to the property of the task.
We use the same U-net architecture as SeCo for the change detection task, which employ the pre-trained backbone models to extract features as the encoder in the U-net.
We apply the pre-trained WRN-50-2 and ResNet-50 backbone models as the encoder and select the first convolution layer, and Layer 1 to 4 as the copy and cropping layers.
As for the decoder module, it is  constructed  to rebuild an image with the same width and height as the input.
And the input and output sizes of its layers are set according to the input and output of the specific encoder layers, in order to concatenate them in the direction of the channel.
Upsampling here is achieved by interpolate function, which can be understood simply as a technology to increase the resolution of the output.
The last layer of the U-net is a $1\times 1$ convolutional layer, which is used to map the channel of the feature vector to the number of output classes.
During fine-tuning, in order to avoid overfitting, only the U-net weights are updated.
In this task, the pre-trained WRN are fine-tuned with a batch size of $32$ and a learning rate of $0.0006$.
The loss function used for this task is $BCEWithLogitsLoss + SoftDiceLoss$.

\textbf{Quantitative Results}
\cref{table:OSCD_results} provides the results of DINO, DINO-MC, and DINO-TC with ResNet-50 and WRN-50-2 respectively, compared against some supervised and self-supervised baselines on OSCD dataset.
The F1 score of DINO-MC with ResNet-50 is $5.52\%$ higher than that of SeCo and almost $3\%$ higher than that of DINO.
When using WRN-50-2 as the backbone, the F1 score of DINO-MC is $5.76\%$ higher than that of SeCo and similar with DINO.
It is no surprise that DINO-TP does not perform as well as DINO and DINO-MC, because DINO-TP receives images of the same location taken at different times as positive instances, so it aims to learn features that do not change over time, which is not very suitable for change detection task \cite{manas2021seasonal}.
In the results of OSCD task, the backbones pre-trained in DINO-MC can capture the subtle differences of image changes over time after a simple fine-tuning on the change detection dataset, which indicates that DINO-MC is able to learn general and effective features using only very simple data augmentation methods.

\textbf{Qualitative Results}
\cref{Figure:OSCD_visualization} gives the visualization masks of DINO-MC on OSCD task.
To do comparison to SeCo, we select two identical examples from the OSCD validation set and present their results.
The masks generated by DINO-MC outperform SeCo-1M since they cover more changed pixels without excessive false predictions.
We observe that the performance of DINO-TP on OSCD task is unstable since it achieves particularly high F1 score on the first instance, which is $4.35$ higher than SeCo-1M, but much lower in the second one, which is $2.25$ lower than SeCo-1M. 
Although positive temporal contrast used in DINO-TP has been proved to be undesirable for change detection task \cite{manas2021seasonal}, the performance of DINO-TP is comparable to SeCo-1M when evaluated on the whole validation set.


\section{Conclusions}
This study introduces DINO-TP and DINO-MC, which expand upon the contrastive SSL framework in two ways: (1) by adding a positive temporal contrast strategy, and (2) by introducing a novel cropping and color transformation strategy for local views.
The results of KNN and linear probing evaluation on EuroSAT demonstrate the effectiveness of our cropping strategy, as well as the unstable performance of the positive temporal contrast.
We evaluate and compare our models with established supervised and self-supervised models on two land use classification tasks and one change detection task in remote sensing domain, revealing the superior performance and effectiveness of DINO-MC.

This study only focus on utilizing temporal views as the positive instances, which does not demonstrate its superiority in our experimental results.
In future work, we will further investigate the performance of applying temporal views as negative instances in the contrastive SSL methods.


\textbf{Acknowledgement}
This research was supported by The University of Melbourne’s Research Computing Services and the Petascale Campus Initiative.

{
    \small
    \bibliographystyle{ieeenat_fullname}
    \bibliography{main}
}


\end{document}